\documentclass{article}
\pdfoutput=1

\usepackage[final]{corl_2019} 

\usepackage{amsmath}
\usepackage{bm}
\usepackage{amssymb}
\usepackage{graphicx}
\usepackage{subcaption}
\usepackage[makeroom]{cancel}
\usepackage[dvipsnames]{xcolor}

\usepackage{authblk}

\usepackage{natbib}
\newcommand{\KL}{\textrm{KL}}
\newcommand{\ELBO}{\textrm{ELBO}}
\DeclareMathOperator{\argmax}{argmax}

\title{Variational Inference for Model-Free and Model-Based Reinforcement Learning}

\author{Felix Leibfried}

\begin{document}
\maketitle


\begin{abstract}
Variational inference (VI) is a specific type of approximate Bayesian inference that approximates an intractable posterior distribution with a tractable one. VI casts the inference problem as an optimization problem, more specifically, the goal is to maximize a lower bound of the logarithm of the marginal likelihood with respect to the parameters of the approximate posterior. Reinforcement learning (RL) on the other hand deals with autonomous agents and how to make them act optimally such as to maximize some notion of expected future cumulative reward. In the non-sequential setting where agents' actions do not have an impact on future states of the environment, RL is covered by contextual bandits and Bayesian optimization. In a proper sequential scenario, however, where agents' actions affect future states, instantaneous rewards need to be carefully traded off against potential long-term rewards. This manuscript shows how the apparently different subjects of VI and RL are linked in two fundamental ways. First, the optimization objective of RL to maximize future cumulative rewards can be recovered via a VI objective under a soft policy constraint in both the non-sequential and the sequential setting. This policy constraint is not just merely artificial but has proven as a useful regularizer in many RL tasks yielding significant improvements in agent performance. And second, in model-based RL where agents aim to learn about the environment they are operating in, the model-learning part can be naturally phrased as an inference problem over the process that governs environment dynamics. We are going to distinguish between two scenarios for the latter: VI when environment states are fully observable by the agent and VI when they are only partially observable through an observation distribution.
\end{abstract}

\keywords{Variational Inference, Contextual Bandits, Model-Free Reinforcement Learning, Model-Based Reinforcement Learning}




\section{Introduction}
\label{sec:intro}

Computing an exact posterior distribution according to Bayes' rule is intractable for many likelihood-prior pairs. The reason is that the marginal likelihood, which is the denominator in Bayes' rule, requires to integrate over the inferred variable and this integral is often intractable. VI translates the problem of Bayesian inference into an optimization objective, namely to maximize a lower bound of the logarithm of the marginal likelihood with respect to the parameters of an approximate but tractable posterior. The log marginal likelihood is a popular objective for identifying optimal parameters of the generative model in a wide range of inference tasks~\citep{Bishop2006}. Since the VI objective poses a lower bound to the log marginal likelihood, it enables hence convenient joint optimization over variational parameters (that parameterize the approximate posterior) and generative parameters (that describe the generative model i.e.\ the prior and the likelihood) at the same time.

RL, where agents aim at maximizing expected future cumulative reward, can be studied in a non-sequential and a sequential setting. In the non-sequential setting, an agent's actions do not affect future states of the environment while in the sequential setting they do. This is covered by contextual bandits (or Bayesian optimization in case of continuous actions) and actual RL respectively. It has been practically proven useful to constrain an agent's optimal policy to be close to a reference policy trough a Kullback-Leibler constraint resulting in improved agent performance (see e.g.\ \cite{Haarnoja2018}).
In this paper, we show how to recover said constraint from a VI angle in line with \cite{Levine2018} but we also show how to use VI to learn environment models for model-based RL.

The remainder of this manuscript is organized as follows. In Section~\ref{sec:vi}, we provide a general background over VI. In Section~\ref{sec:contextual_bandits}, we demonstrate how one can recover the non-sequential RL objective under a soft policy constraint when starting from a VI formulation. In Section~\ref{sec:rl}, we do the same but for a sequential scenario that is typical for most of RL. In Section~\ref{sec:mbrl}, we show how to leverage VI for model-based RL in order to infer the dynamics of the environment which is unknown to the agent. In Section~\ref{sec:mbrl_po}, we show how to infer environment dynamics with VI for a partially observable setting where agents cannot observe environment states directly but only through a noisy observation distribution. In Section~\ref{sec:conclusion}, we conclude with a summary.

\section{Variational Inference}
\label{sec:vi}

The goal of this section is to provide an overview of VI, where we resort mostly to the parameter space view and where we assume a supervised learning scenario. Note that this section is in large parts identically overlapping with Section~3.1 from~\cite{Leibfried2020} which contained the most essential bits of this paper in an earlier version but we decided to ``pull out'' the RL bits later on.

\begin{figure}[h!]
\centering
\includegraphics[trim=0 230 420 40,clip,width=0.6\textwidth]{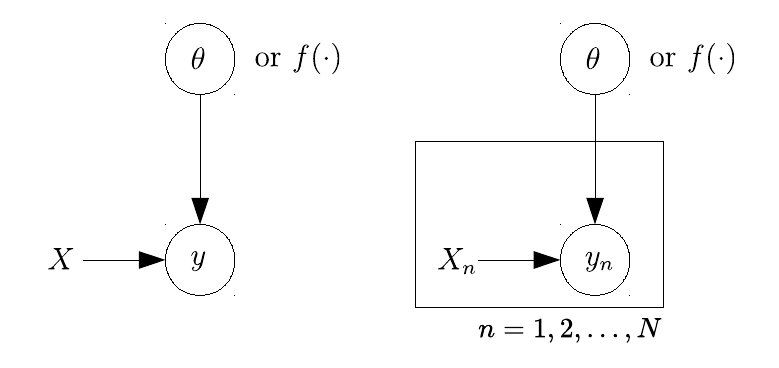}\caption{Graphical models for VI in supervised learning settings. $X$ refers to inputs and $y$ to labels. Unknown functions are depicted via the variable $\theta$ (in parameter space view) or $f(\cdot)$ (in function space view). We present one general formulation and another one in which the data set is i.i.d.\ which is a common assumption in most scenarios---individual training examples are then indexed with $n$.}
\label{fig:vi}
\end{figure}

Let's start be revisiting Bayesian inference for supervised learning. Imagine some input $X$, some observed variable $y$ and the probability of observing $y$ given $X$ via a parametric distribution $p_\gamma(y | \theta, X)$ with hyperparameters $\gamma$ and unknown parameters $\theta$. Our goal is to infer $\theta$ and we have some prior belief over $\theta$ through the distribution $p_\gamma(\theta)$, where we assume for notational convenience that both $p_\gamma(y | \theta, X)$ and $p_\gamma(\theta)$ are hyperparameterized by $\gamma$. Inference over $\theta$ is then obtained via the posterior distribution over $\theta$ after observing $y$ and $X$:
\begin{equation}
\label{eq:bayes_rule}
p_\gamma(\theta|y, X) = \frac{p_\gamma(y | \theta, X) p_\gamma(\theta)}{\int p_\gamma(y | \theta, X) p_\gamma(\theta) \mathrm{d}\theta},
\end{equation}
where $p_\gamma(y | \theta, X)$ is referred to as likelihood, $p_\gamma(\theta)$ as prior and $p_\gamma(y|X) = \int p_\gamma(y | \theta, X) p_\gamma(\theta) \mathrm{d}\theta$ as marginal likelihood (or evidence). The corresponding graphical model for this inference problem is depicted in Figure~\ref{fig:vi} on the left-hand side (denoted as ``general formulation''). The graphical model represents the joint distribution $p_\gamma(y | \theta, X) p_\gamma(\theta)$ between $y$ and $\theta$ given $X$, which is also referred to as ``generative model''. The challenge in computing the posterior is that the marginal likelihood usually does not have a closed form solution except for special cases, e.g.\ when the prior is conjugate to the likelihood (which we won't consider in this tutorial). When the marginal likelihood does have a closed form solution, it is usually maximized w.r.t.\ to the hyperparemeters~$\gamma$ of the generative model before the exact posterior is computed~\citep{Bishop2006,Rasmussen2006}. Note that the hyperparameters $\gamma$ are also referred to as ``generative parameters''.

The idea in VI is to introduce an approximation $q_\psi(\theta)$, parameterized via $\psi$, to the intractable posterior $p_\gamma(\theta|y, X)$, and to optimize for $\psi$ such that the approximate posterior becomes close to the true posterior. In this regard, the approximate posterior $q_\psi(\theta)$ is also referred to as ``variational model'' and $\psi$ as ``variational parameters''. The question is which optimization objective to choose to identify optimal variational parameters $\psi$. We are going to respond to this question shortly but for now, we commence with the negative Kullback-Leibler divergence ($\KL$) between the approximate and the true posterior, which can be written as (by applying Bayes' rule to the true posterior):
\begin{equation}
\label{eq:neg_kl}
-\KL\Big(q_\psi(\theta) \Big|\Big| p_\gamma(\theta|y, X) \Big) = \int q_\psi(\theta) \ln p_\gamma(y | \theta, X) \; \mathrm{d}\theta -\KL\Big(q_\psi(\theta) \Big|\Big| p_\gamma(\theta) \Big) - \ln p_\gamma(y|X) .
\end{equation}
Rearranging by bringing the log marginal likelihood term $\ln p_\gamma(y|X)$ to the left yields:
\begin{equation}
\label{eq:elbo}
\ln p_\gamma(y|X) - \KL\Big(q_\psi(\theta) \Big|\Big| p_\gamma(\theta|y, X) \Big) = \underbrace{\int q_\psi(\theta) \ln p_\gamma(y | \theta, X) \; \mathrm{d}\theta -\KL\Big(q_\psi(\theta) \Big|\Big| p_\gamma(\theta) \Big)}_{=:\ELBO(\gamma,\psi)}.
\end{equation}
The term on the right-hand side is referred to as the evidence lower bound $\ELBO(\gamma,\psi)$~\citep{Rasmussen2006} since it poses a lower bound to the log marginal likelihood (a.k.a.\ log evidence)---``log evidence lower bound'' might hence be a more suitable description but omitting ``log'' is established convention. The $\ELBO$ is a lower bound because the $\KL$ between the posterior approximation and the true posterior is non-negative. Since the log marginal likelihood does not depend on the variational parameters $\psi$, the $\ELBO$ assumes its maximum when the approximate posterior equals the true one, i.e.\ $q_\psi(\theta) = p_\gamma(\theta|y, X)$, in which case the $\KL$ term on the left is zero and the $\ELBO$ recovers the log marginal likelihood exactly. 

Note how the formulation for the $\ELBO$ does not require to know the true posterior in its functional form a priori in order to identify an optimal approximation, because Equation~\eqref{eq:elbo} was obtained via decomposing the intractable posterior via Bayes' rule. Also note that the log marginal likelihood is usually the preferred objective to maximize for the generative hyperparameters $\gamma$, as mentioned earlier. Contemporary VI methods therefore maximize the evidence lower bound $\max_{\gamma,\psi}\ELBO(\gamma,\psi)$ jointly w.r.t.\ both generative parameters $\gamma$ and variational parameters $\psi$. Some current methods with deep function approximators choose a slight modification of Equation~\eqref{eq:elbo} by multiplying the $\KL$ term between the approximate posterior and the prior with a positive $\beta$-parameter. This is called ``$\beta$-VI'' and recovers a maximum expected log likelihood objective as a special case when $\beta \rightarrow 0$. It has been proposed by~\cite{Higgins2017}, and \cite{Wenzel2020} provide a recent discussion.

Assuming an optimal approximate posterior has been identified after optimizing the $\ELBO$ w.r.t.\ both variational parameters $\psi$ and generative parameters $\gamma$, the next question is how to use it, namely how to predict $y^\star$ for a new data point $X^\star$ that is not part of the training data. The answer is:
\begin{equation}
\label{eq:pred}
p(y^\star | X^\star) =  \int p_\gamma(y^\star | \theta, X^\star) q_\psi(\theta) \; \mathrm{d}\theta ,
\end{equation}
by forming the joint between the likelihood $p_\gamma(y^\star | \theta, X^\star)$ and the approximate posterior $q_\psi(\theta)$, and integrating out $\theta$. If the integration has no closed form, one has to resort to Monte Carlo methods---i.e.\ replace the integral over $\theta$ with an empirical average via samples obtained from $q_\psi(\theta)$.

So far, we haven't made any assumptions on how the generative model looks like precisely. In supervised learning, it is however common to assume an i.i.d.\ dataset in the sense that the training set comprises $N$ i.i.d.\ training examples in the form of ($X_n, y_n$)-pairs. The corresponding graphical model is depicted in Figure~\ref{fig:vi} on the right denoted as ``i.i.d.\ dataset''. In this case, the likelihood is given by $\prod_{n=1}^N p_\gamma(y_n | \theta, X_n)$ and the $\ELBO$ becomes:
\begin{equation}
\label{eq:elbo_iid}
\ELBO(\gamma,\psi) = \sum_{n=1}^N \int q_\psi(\theta) \ln p_\gamma(y_n | \theta, X_n) \; \mathrm{d}\theta -\KL\Big(q_\psi(\theta) \Big|\Big| p_\gamma(\theta) \Big).
\end{equation}
An interesting fact to note is that in case of large $N$, Equation~\eqref{eq:elbo_iid} can be approximated by Monte Carlo using minibatches, which can be used for parameter updates without exceeding potential memory limits~\citep{Hensman2013}.
The corresponding predictions $\{y_n^\star\}_{n=1,..,N^\star}$ for new data points $\{X_n^\star\}_{n=1,..,N^\star}$ in the i.i.d.\ setting are:
\begin{equation}
\label{eq:pred_iid}
p(y_1^\star, ..., y_{N^\star}^\star | X_1^\star, ..., X_{N^\star}^\star) =  \int \prod_{n=1}^{N^\star} p_\gamma(y^\star_n | \theta, X^\star_n) q_\psi(\theta) \; \mathrm{d}\theta .
\end{equation}

Note that we have deliberately not made any assumptions on the dimensions of $y$, $X$, $\theta$, $\gamma$ and $\psi$ to keep the notation light (which doesn't mean that these quantities need to be scalars). We also chose the weight space view by using $\theta$ instead of the function space view, although both are conceptually equivalent. The function space view can be obtained by replacing $\theta$ with $f(\cdot)$ in all formulations and equations above. Practically however, one would need to be careful with expectations and $\KL$ divergences between infinite-dimensional random variables~\citep{Matthews2016}.

\section{Contextual Bandits as Inference}
\label{sec:contextual_bandits}

Contextual bandits cover a class of non-sequential decision-making problems~\citep{Lattimore2020}. Formally, there is a state or context variable $s \in \mathcal{S}$ and an action variable $a \in \mathcal{A}$, and the decision-maker's task is to take in each state $s$ an optimal action $a$ where optimality is defined through some real-valued reward function $R(s,a): \mathcal{S} \times \mathcal{A} \rightarrow \mathbb{R}$. Clearly, the optimal strategy would be $\argmax_a R(s,a)$ if the reward-maximizing argument can be identified for a given state~$s$. Note that a finite action set $\mathcal{A}$ refers to the classical contextual bandit setting while a continuous action set describes the setting of contextual Bayesian optimization. 

\begin{figure}[h!]
\centering
\includegraphics[trim=0 150 0 100,clip,width=1.0\textwidth]{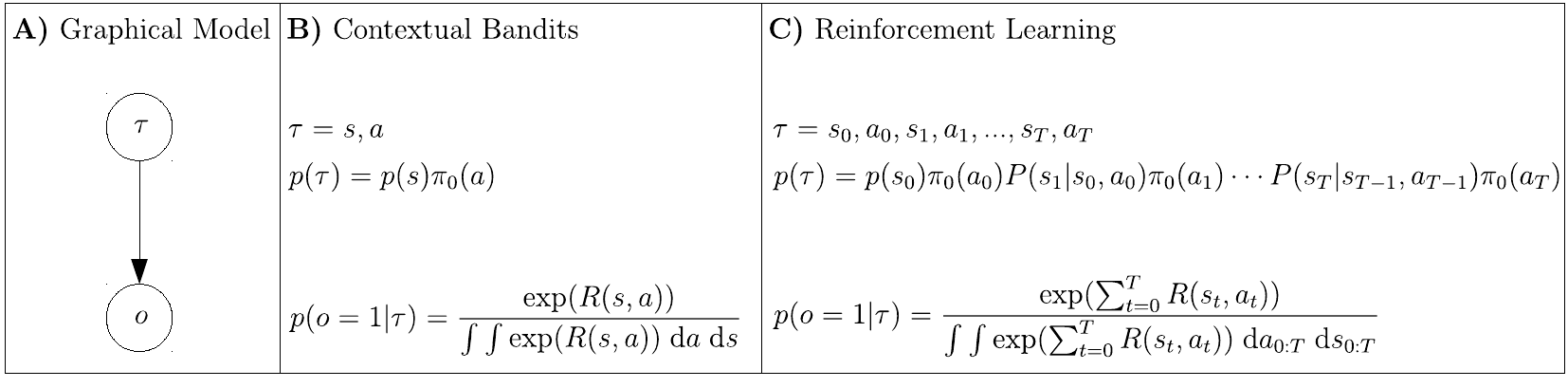}
\caption{Graphical models when interpreting decision-making as inference problems. \textbf{A)} shows a generic graphical model with a latent variable $\tau$ and an observed binary variable $o \in \{0, 1\}$. Both of these variables have similar but different meanings in non-sequential decision-making settings of the contextual bandits type (as depicted in \textbf{B})), and in sequential decision-making settings of the reinforcement learning type (as depicted in \textbf{C})). In the contextual bandits setting (Section~\ref{sec:contextual_bandits}), $\tau$~refers to a state-action pair $(s,a)$ and the probability of $o=1$ conditioned on $\tau$ is proportional to the $\exp$ of the reward $R(s,a)$ obtained when executing the action $a$ in the state $s$. Here, $p(s)$ refers to the distribution over states and $\pi_0(a)$ to a prior action distribution. In the reinforcement learning setting (Section~\ref{sec:rl}), $\tau$ refers to a state-action trajectory $(s_0,a_0,s_1,a_1,...,s_T,a_T)$ within a Markov decision process, and the probability of $o=1$ conditioned on $\tau$ is proportional to the $\exp$ of the cumulative reward $\sum_{t=0}^T R(s_t,a_t)$. For the latter, we assume a finite time horizon where $T$ indicates the end of the horizon. The notions $s_{0:T}$ and $a_{0:T}$ are short-hand notations for state and action trajectories respectively. $p(s_0)$ refers to the initial state distribution at time step $t=0$, $\pi_0(a_t)$ to a prior action distribution as in the non-sequential setting, and $P(s_{t+1}|s_t,a_t)$ refers to the state-transition probability distribution of transitioning to the state $s_{t+1}$ at time step $t+1$ when executing the action $a_t$ in the state $s_t$ at time step $t$. Note that in both decision-making settings, the assumption is always that there is only one single observation for $o$ which is $1$---see text for details.}
\label{fig:vi_cb_rl}
\end{figure}

It turns out that this simple decision-making paradigm can be recovered under a soft policy constraint from an inference formulation, as explained in the following. For convenience, let's introduce the latent variable $\tau = (s,a)$ as short-hand notation for a state-action pair. Imagine a factorized prior for $\tau$ of the form $p(\tau) = p(s) \pi_0(a)$, where $p(s)$ is a prior distribution over states and $\pi_0(a)$ is a prior action distribution. Phrasing an inference problem requires then to specify an observed variable $o$ and a likelihood that determines the probability of $o$ conditioned on the latent variable $\tau$. We make the artificial choice of $o \in \{0, 1\}$ being binary and define $p(o=1|\tau) = \frac{\exp(R(s,a))}{\int \int \exp(R(s,a)) \textrm{d} a \textrm{d} s}$. We furthermore artificially determine that only one observation $o$ has been made and that this observation is $1$. The corresponding graphical model plus a description of the prior and the likelihood is depicted in Figure~\ref{fig:vi_cb_rl}~A) and~B). Under an approximate posterior of the form $q(\tau) = p(s) \pi(a|s)$ that uses a state-conditioned action distribution $\pi(a|s)$, a.k.a.\ policy, we arrive at the following $\ELBO$:
\begin{eqnarray}
\ELBO(\pi_0, \pi) &=& \int q(\tau) \ln p(o=1 | \tau) \; \mathrm{d} \tau  - \KL\Big(q(\tau) \Big|\Big| p(\tau) \Big) \label{eq:elbo_cb} \\
&\equiv& \int \int p(s) \pi(a|s) R(s,a) \; \textrm{d} a \; \textrm{d} s \; - \int p(s)  \KL \Big( \pi(a|s) \Big|\Big| \pi_0(a)\Big) \; \textrm{d} s, \label{eq:elbo_cb2}
\end{eqnarray}
where the equivalence sign ``$\equiv$'' in Equation~\eqref{eq:elbo_cb2} is due to the fact that the normalization constant in the denominator of $p(o=1|\tau)$ does neither depend on the generative action prior $\pi_0$ nor on the variational policy $\pi$, and is hence dropped since it would just add an additional term that is unaffected by the optimization arguments $(\pi_0, \pi)$.

It can be seen that the objective in Equation~\eqref{eq:elbo_cb2} recovers the original contextual bandit problem when not optimizing over the prior $\pi_0$ but only over the policy $\pi$, but importantly under a soft policy constraint that encourages the optimal policy to be close to the prior action distribution. In fact, one can weight the expected $\KL$ penalty in Equation~\eqref{eq:elbo_cb2} with a non-negative scalar $\beta$ to control the trade-off between reward maximization and penalization---we touched on this approach before in Section~\ref{sec:vi} where it was referred to as $\beta$-VI\@. In $\beta$-VI, one recovers the contextual bandit objective precisely as $\beta \rightarrow 0$. The decision-making counterpart of $\beta$-VI is called ``information-theoretic bounded rationality'', see~\cite{Ortega2013} and~\cite{Ortega2015} for details.

Going one step further, one can optimize over the generative action prior $\pi_0$ in addition to the variational policy $\pi$. It turns out that the optimization problem becomes then mathematically equivalent to rate distortion theory from information theory~\citep{Cover2006}. In short, under the latter view, the decision-maker can be interpreted as information-theoretic channel with input $s$ and output $a$, the reward function as some measure of quality for recovering the information of $s$ in $a$, and the $\KL$ penalty as maximal information transmission rate. A detailed discussion of the topic is however beyond the scope of this tutorial and we refer the interested reader to \cite{Genewein2015} who provide an in-depth discussion. 

More concrete applications of the rate distortion framework for decision-making can be found in the literature across different research fields, e.g.\ economics~\citep{Sims2010}, theoretical neuroscience~\citep{Leibfried2015} and machine learning~\citep{Leibfried2016}. Most recent advances aim at extending the framework to hierarchical decision-making settings, where lower-level decision-making routines pre-process the context yielding intermediate outcomes, that are then further processed by higher-level decision-making routines~\citep{Peng2017,Hihn2019,Hihn2020}.

\section{Reinforcement Learning as Inference}
\label{sec:rl}

As opposed to contextual bandits, reinforcement learning~\citep{Sutton1998} covers sequential decision-making problems where an agent needs to identify a sequence of actions that maximizes cumulative reward.
Mathematically speaking, the agent operates in an environment that is in a certain state $s_t \in \mathcal{S}$ at every time step $t$. The agent needs to choose an action $a_t \in \mathcal{A}$ after observing~$s_t$. This yields an instantaneous reward $R(s_t, a_t): \mathcal{S} \times \mathcal{A} \rightarrow \mathbb{R}$ as defined by a real-valued reward function (as in the contextual bandit case), and the environment transitions to a subsequent state~$s_{t+1}$. Importantly, transitioning to a subsequent state is determined by a state-transition probability distribution $P(s_{t+1}|s_t,a_t)$ that is a conditional probability distribution over $s_{t+1}$ conditioned on $s_t$ and $a_t$. This problem definition describes a Markov decision process (MDP) and the agent's task is to maximize expected cumulative reward $\mathbb{E} \Big[ \sum_{t=0}^{T} R(s_t,a_t) \Big]$ where $T$ determines the time horizon. Here, the expectation is w.r.t.\ three components: the initial state distribution $p(s_0)$ that samples the first state at time step $t=0$, the state-transition distribution $P(s_{t+1}|s_t,a_t)$, and the agent's policy $\pi(a_t|s_t)$ which is a conditional probability distribution over $a_t$ conditioned on $s_t$. The goal of the agent is hence to find an optimal policy $\pi(a_t|s_t)$ that yields highest expected cumulative reward. 

The difficulty compared to the contextual bandit setting is that an action $a_t$ at time step $t$ not only affects the instantaneous reward at that time step, but also future rewards through future states.
Note that we chose a finite-horizon problem with a stationary policy for the problem definition, because it will come in handy shortly when expressing the reinforcement learning problem as an inference problem. Also note that the contextual bandit formulation from the previous section can be recovered as a special case when defining the initial state distribution $p(s_0)$ to be the same as the state-transition distribution, i.e.\ $P(s_{t+1}|s_t,a_t) = p(s_{t+1})$ which does not depend on $s_t$ and $a_t$.

Phrasing the reinforcement learning problem as an inference problem is conceptually similar to the contextual bandit scenario, but slightly more complex due to the sequential nature of the problem. In line with the previous section, let's introduce a latent variable $\tau = s_0,a_0,s_1,a_1,...,s_T,a_T$ as short-hand notation for a state-action trajectory. Assume a factorized prior for $\tau$ of the following form: $p(\tau) = p(s_0) \pi_0(a_0) \prod_{t=0}^{T-1} P(s_{t+1}|s_t,a_t) \pi_0(a_{t+1})$ where $\pi_0(a_t)$ refers to a prior action distribution. We again introduce an artificial binary observed variable $o \in \{0,1\}$ and define the likelihood as $p(o=1|\tau) = \frac{\exp(\sum_{t=0}^T R(s_t,a_t))}{\int \int \exp(\sum_{t=0}^T R(s_t,a_t)) \mathrm{d} a_{0:T} \mathrm{d} s_{0:T}}$ where $s_{0:T}$ and $a_{0:T}$ is short-hand for state- and action-trajectories respectively. As earlier, we assume that only one observation~$o$ has been made and that this observation is $1$. This is summarized in Figure~\ref{fig:vi_cb_rl}~A) and~C). Using an approximate posterior of the form $q(\tau) = p(s_0) \pi(a_0|s_0) \prod_{t=0}^{T-1} P(s_{t+1}|s_t,a_t) \pi(a_{t+1}|s_{t+1})$ with a state-conditioned variational policy $\pi(a_t|s_t)$, we obtain an $\ELBO$ expression as:
\begin{eqnarray}
\ELBO(\pi_0, \pi) &=& \int q(\tau) \ln p(o=1 | \tau) \; \mathrm{d} \tau  - \KL\Big(q(\tau) \Big|\Big| p(\tau) \Big) \label{eq:elbo_rl} \\
&\equiv& \sum_{t=0}^T \int \int p_t^{(\pi)}(s_t) \pi(a_t|s_t) R(s_t,a_t) \; \textrm{d} a_t \; \textrm{d} s_t \nonumber \\
&&- \sum_{t=0}^T \int p_t^{(\pi)}(s_t)  \KL \Big( \pi(a_t|s_t) \Big|\Big| \pi_0(a_t)\Big) \; \textrm{d} s_t, \label{eq:elbo_rl2}
\end{eqnarray}
where $p_t^{(\pi)}(s_t)$ refers to the marginal state distribution of the policy $\pi$ at time step $t$, which is obtained by taking $q(\tau)$ up to $t$ and integrating out all trajectory elements except for the state at the last time step~$t$. As in the previous section, the equivalence sign ``$\equiv$'' in Equation~\eqref{eq:elbo_rl2} is because the denominator in $p(o=1|\tau)$ can be dropped since it just leads to an additive term that does not depend on $\pi_0$ and $\pi$.

It becomes apparent that Equation~\eqref{eq:elbo_rl2} recovers the original reinforcement learning objective but under a soft policy constraint that encourages the optimal policy to be close to a prior action distribution (when optimizing for $\pi$ but keeping $\pi_0$ fixed). A more detailed derivation can be found in \cite{Levine2018}, but the underlying problem description remains also valid when using a state-conditioned action prior $\pi_0(a_t|s_t)$ rather than a state-unconditioned one. Similar to before, we can introduce a non-negative scaling factor $\beta$ to weight the $\KL$ penalty at each time step. The latter can be referred to as ``soft cumulative reward maximization'' which has been described in infinite-horizon MDP settings~\citep{Azar2011,Rubin2012,Neu2017} where $T \rightarrow \infty$.

Under the special case of using a uniform prior action distribution $\pi_0(a_t)=\mathcal{U}(\mathcal{A})$, one arrives at the framework of entropy regularization for MDPs~\citep{Ziebart2010,Nachum2017}. Entropy regularization for MDPs has spurred quite a few reinforcement learning algorithms in the recent years that improved performance over ordinary non-regularized algorithms in a wide range of domains. These domains include grid worlds~\citep{Fox2016}, robotics domains~\citep{Haarnoja2017,Haarnoja2018,Haarnoja2019}, but also visual domains such as Atari games~\citep{Schulman2017,Leibfried2018}. In the meanwhile, the concept found also application in two-player scenarios where it is referred to as ``tuneable artificial intelligence'' meant to balance an artificial player's performance through the $\KL$ penalty such as to keep a human player engaged---see~\cite{Grau2018} for details.

When optimizing over $\pi_0$ in addition to $\pi$, a generalization of the rate distortion framework for non-sequential decision-making (presented at the end of the previous section) is obtained. This was first described by~\cite{Tishby2011}, and recent advances led to reinforcement learning algorithms that improved over entropy-regularized methods in Atari and robotics domains~\citep{Grau-Moya2019,Leibfried2019}. While the rate distortion framework fits into the inference perspective, recent research presents alternative information-theoretic formulations beyond rate distortion theory which enable more sophisticated prior action distributions~\citep{Tiomkin2018,Leibfried2019b} that can be conditioned on past or future observations.

We conclude this section by pointing out that $\KL$ regularization is a widely adopted technique in various fields of sequential decision-making. While enumerating all of them here would be beyond the scope of this manuscript, we want to point out a few in the further vicinity of soft cumulative reward maximization. Notable use of different types of $\KL$ regularization have been adopted in policy search~\citep{Peters2010,Schulman2015} and decision theory~\citep{Karny2020}. But also noteworthy is work in the field of robust MDPs~\citep{Grau-Moya2016} and multi-task learning~\citep{Petangoda2019} where $\KL$ regularization is applied on the level of environment models or latent-variable distributions respectively (in addition to the action level).

\section{Variational Inference for Model-Based Reinforcement Learning}
\label{sec:mbrl}

The reinforcement-learning-as-inference framework usually assumes a ``model-free'' perspective where an agent learns by trial and error, but does neither know nor try to infer an environment model. An environment model is useful for ``model-based'' reinforcement learning that enables an agent e.g.\ to virtually plan ahead in order to identify an optimal course of action. The prospect of model-based learning is to be more sample-efficient, i.e.\ to obtain good performance with less agent-environment interactions, but the drawback is that additional computational resources are required both for model learning but also for planning (which requires an internal virtual simulation). Since model-based learning has received increased attention in the recent years, we decided to dedicate a separate section to the subject in order to outline how to use VI to learn an environment model from agent-environment interaction data. Note that a tutorial on model-based reinforcement learning is beyond the scope of this manuscript, but we will provide some examples at the end of this section to provide an intuition for how models can be used for sequential decision-making.

\begin{figure}[h!]
\centering
\includegraphics[trim=210 130 260 150,clip,width=0.5\textwidth]{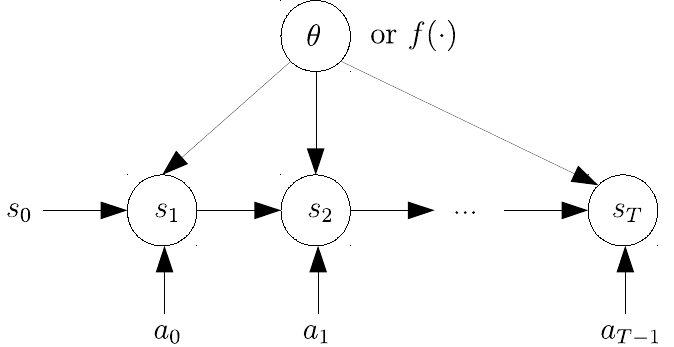}
\caption{Graphical model for VI to infer the state-transition distribution for model-based reinforcement learning. Environment states are indicated with $s_t$ and actions, taken by the agent, with $a_t$ where $t \in \{0,1,...,T\}$ is a time index. The unknown state-transition distribution is denoted as $\theta$ in parameter space view (or as $f(\cdot)$ in function space view). Note that the inference procedure is conditioned on the initial state $s_0$ as well as on all actions from $a_0$ up to $a_{T-1}$.}
\label{fig:vi_mbrl}
\end{figure}

Going ahead to phrase an inference problem, we need to specify a generative model first. For the sake of simplicity, we assume in the following that we are only interested in inferring the state-transition distribution, but not the initial state distribution or the reward function (although this would be possible with straightforward extensions). For this reason, our generative model is conditioned on the initial state and all actions. This leads us to the graphical model in Figure~\ref{fig:vi_mbrl}. The basic idea is that the state-transition distribution $P(s_{t+1}|s_t,a_t)$ from the previous section is expressed through some function with parameters $\theta$ as $P_\theta(s_{t+1}|s_t,a_t)$. We can best imagine $\theta$ as a function that maps a state-action pair $(s_t,a_t)$ to some output that then determines the distribution over $s_{t+1}$. Hence, we could have similarly expressed this function in function space view as $f(\cdot)$, where the dot $\cdot$ refers to the concatenated state-action domain, and the state-transition distribution as $P_{f(\cdot)}(s_{t+1}|s_t,a_t)$. 

However, such function space transitions models are more difficult to sample from compared to ordinary weight space models. The reason is that state-transition models are auto-regressive, i.e.\ their output serves as input for the next time step, and it is not possible to sample a function from a process a priori without knowing where to evaluate that function (as opposed to weight space models). We refer the interested reader to \cite{Wilson2020} which could provide a remedy.

Under the former assumption, we need to determine a prior distribution over $\theta$ as $p_\gamma(\theta)$ with generative parameters $\gamma$. The likelihood is then simply defined as $p(s_1,...,s_T|s_0,a_0,...a_{T-1},\theta) = \prod_{t=1}^T P_\theta(s_{t+1}|s_t,a_t)$. All that is left is the specification of an approximate posterior $q_\psi(\theta)$ with variational parameters $\psi$ in order to phrase the subsequent $\ELBO$:
\begin{eqnarray}
\ELBO(\gamma, \psi) &=& \int q_\psi(\theta) \ln p(s_1,...,s_T|s_0,a_0,...a_{T-1},\theta) \; \textrm{d} \theta - \KL \Big( q_\psi(\theta) \Big| \Big| p_\gamma(\theta)  \Big) \label{eq:vi_mbrl} \\
&=& \sum_{t=0}^{T-1} \int q_\psi(\theta) \ln P_\theta(s_{t+1}|s_t,a_t) \; \textrm{d} \theta - \KL \Big( q_\psi(\theta) \Big| \Big| p_\gamma(\theta)  \Big). \label{eq:vi_mbrl_2}
\end{eqnarray}
Note that we could have straightforwardly adopted a function space view by modeling the function prior $p_\gamma(f(\cdot))$ and approximate posterior $q_\psi(f(\cdot))$ with distributions over functions. This would require to swap $P_\theta(s_{t+1}|s_t,a_t)$ with $P_{f(\cdot)}(s_{t+1}|s_t,a_t)$ in Equations~\eqref{eq:vi_mbrl} and~\eqref{eq:vi_mbrl_2}. Also note that due to the i.i.d.\ nature of the likelihood, it is not required to keep track of the temporal order of state-action-next-state tuples $(s_t, a_t, s_{t+1})$. It is in fact only required that each individual state-action-next-state tuple is temporally consistent, but the entirety of tuples is not required to represent a temporally consistent state-action trajectory. And since the entire procedure is action-conditioned, one also does not need to worry about how agent-environment interactions for training have been obtained, i.e.\ whether from a single agent or multiple agents with different policies does not matter.

Predicting future states $s_1^\star, s_2^\star, ..., s_{T^\star}^\star$ is then accomplished in an initial-state and action-conditioned fashion, given a single initial state $s_0^\star$ and a sequence of actions $a_0^\star, a_1^\star, a_2^\star, ..., a_{T^\star-1}^\star$:
\begin{equation}
p(s_1^\star, ..., s_{T^\star}^\star | s_0^\star, a_0^\star, ..., a_{T^\star-1}^\star) = \int \prod_{t=0}^{T^\star-1} P_\theta(s_{t+1}^\star|s_t^\star,a_t^\star) q_\psi(\theta) \; \mathrm{d} \theta ,
\label{eq:state_prediction}
\end{equation}
where, importantly, the outer expectation is w.r.t.\ $\theta$. This is practically crucial since a future state sequence can be generated by sampling $\theta$ from the approximate posterior $q_\psi(\theta)$ first, and then ``unrolling'' the state-transition model with the sampled $\theta$. More concretely, given $s_0^\star$ and $a_0^\star$, one can sample~$s_1^\star$ from $P_\theta(s_1^\star | s_0^\star, a_0^\star)$. Next, one can sample $s_2^\star$ from $P_\theta(s_2^\star | s_1^\star, a_1^\star)$ that depends on the previous state sample $s_1^\star$, and so forth, until one arrives at the final state~$s_{T^\star}^\star$.

We conclude this section with a brief overview over the current state of the art in model-based reinforcement learning with a particular focus on robotics simulation domains because they provide a challenging benchmark (e.g.\ containing high-dimensional environments with highly non-linear state-transition distributions). We need to emphasize that a vast majority of the mentioned algorithms does not use approximate inference, but non-inference-based objectives in order to identify an environment model. However, since the model-learning component is a modular component of all model-based algorithms, we believe it could be straightforwardly replaced with a proper inference method. While conceptually more principled than non-inference objectives, it remains to be empirically validated if inference-based environment models yield better agent performance.

Most model-based reinforcement learning algorithms can be assigned to one of two categories: decision-time planning and background planning algorithms. Decision-time planners solve a virtual planning problem at every time step in order to identify a reasonable action, while background planners essentially comprise the vast majority of all other model-based algorithms that are not decision-time planners. Contemporary decision-time planners identify a suitable action for a given real environment state by generating several virtual state-action trajectories with the internal environment model starting at the observed state. They then evaluate all of these trajectories regarding their cumulative reward. Subsequently, a trajectory is selected proportional to its cumulative reward, i.e.\ the higher the cumulative reward of a trajectory the higher the probability of being chosen. Finally, the first action of the chosen trajectory is executed in the real environment. To generate virtual trajectories, an internal virtual policy is required. This is where contemporary algorithms divide into two categories: methods with state-unconditioned virtual policies~\citep{Chua2018,Nagabandi2018} and such with state-conditioned ones~\citep{Piche2019,Wang2019}.

While background planning comprises a large class of different algorithms, we focus here on such that train model-free algorithms virtually inside of an environment model. This idea is inspired by the ``DYNA'' architecture presented first in the work of \cite{Sutton1990} for grid world scenarios. In contemporary model-based reinforcement learning algorithms for robotics simulation, two categories of background planners can be found. One class trains ``on-policy'' model-free algorithms virtually inside the environment model~\citep{Kurutach2018,Luo2019} while the other trains ``off-policy'' model-free algorithms virtually~\citep{Janner2019}. The terms ``on-policy'' and ``off-policy'' are standard in the reinforcement learning literature and an explanation can be found in \cite{Sutton1998}. In short, the difference between on-policy and off-policy is that the former needs to re-sample state-action trajectories at a high frequency and cannot naively make use of past trajectory samples, while the latter uses a replay buffer to utilize past trajectory samples at the prize of a possibly higher cumulative reward estimation bias. For a modern software framework for model-based RL that leverages the differences between decision and background planning, see~\cite{McLeod2021}.
For a broader overview of model-based algorithms for robotics simulation, see~\cite{Wang2019b}. Noteworthy mentioning is the work of \cite{Deisenroth2011} and \cite{Kamthe2018} that use Gaussian processes and a proper inference objective for model learning.
There has also been recent progress for model-based reinforcement learning in visual domains~\citep{Leibfried2018b,Kaiser2020,Schrittwieser2020}.

\section{VI for Partially Observable Model-Based Reinforcement Learning}
\label{sec:mbrl_po}

The previous section assumes that environment states~$s_t \in \mathcal{S}$ can be fully observed by the agent. However, in real-world scenarios, states are usually not fully observable. Instead, the agent needs to rely on noisy observations~$o_t \in \mathcal{O}$ from which states~$s_t$, that are in fact latent variables, need to be inferred. Often, $o_t$ does not carry enough information to infer the latent state~$s_t$ precisely. This scenario is commonly referred to as ``partially observable''.

Mathematically, the partially observable setting follows the fully observable setting as described at the beginning of Section~\ref{sec:rl} but requires to extend the MDP definition by observations $o_t$. This is obtained via an observation distribution $P(o_t | s_t)$ that determines how observations~$o_t$ are generated from latent states~$s_t$ through a conditional probability distribution, yielding a partially observable MDP (POMDP) according to~\cite{Kaelbling1998}. Note that the original POMDP definition specifies the observation distribution as $P(o_t | s_t, a_{t-1})$ with an additional dependence on the previous action~$a_{t-1}$ (from which we refrain in this tutorial in order to ease the exposition).

\begin{figure}[h!]
\centering
\includegraphics[trim=380 70 90 60,clip,width=0.5\textwidth]{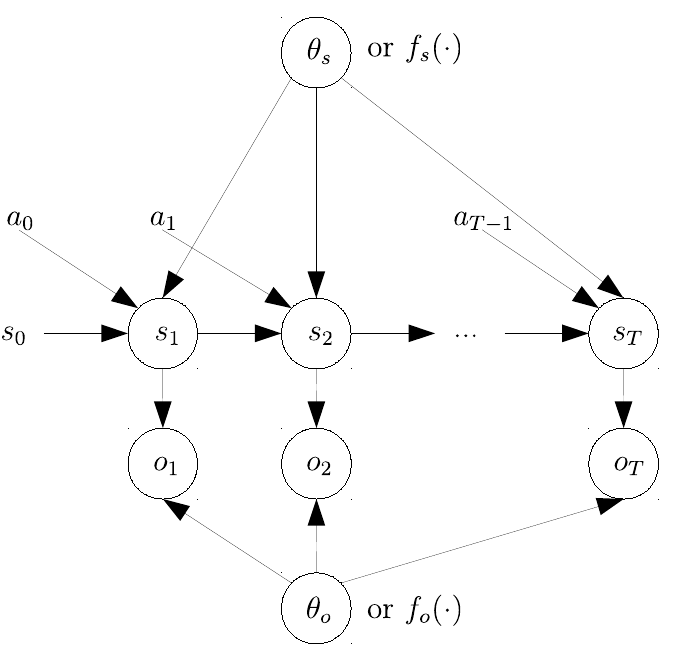}
\caption{Graphical model for VI to infer the state-transition and observation distribution for model-based reinforcement learning in a partially observable setting. Environment states~$s_t$ and actions~$a_t$ are defined the same way as in the fully observable setting from Figure~\ref{fig:vi_mbrl} with a time index $t \in \{0,1,...,T\}$. However, states $s_t$ cannot be directly observed by the agent that needs to rely on noisy observations indicated by $o_t$. The unknown state-transition distribution is denoted as $\theta_s$ in parameter space view (or as $f_s(\cdot)$ in function space view), while the unknown observation distribution is denoted as $\theta_o$ or $f_o(\cdot)$ (in parameter space and function space view respectively). As in the fully observable case, inference is conditioned on the initial state $s_0$ as well as on all actions from $a_0$ up to $a_{T-1}$. Note that if the observation nodes $o_t$ and the observation distribution node $\theta_o$ are removed, one would recover the graphical model from the fully observable case in Figure~\ref{fig:vi_mbrl}.}
\label{fig:vi_mbrl_po}
\end{figure}

To phrase an inference problem, we need to extend the generative model from the fully observable setting as discussed in the previous section (and depicted in Figure~\ref{fig:vi_mbrl}) by observations $o_t$ and parameters $\theta_o$ for the observation distribution $P_{\theta_o}(o_t|s_t)$. Similar to earlier, $\theta_o$ can be imagined as a function that maps a latent state $s_t$ to some output that determines the distribution over $o_t$. We could have alternatively expressed $\theta_o$ in function space view as $f_o(\cdot)$ and the observation distribution as $P_{f_o(\cdot)}(o_t | s_t)$. In order to distinguish parameters for the observation distribution $P_{\theta_o}(o_t|s_t)$ and state transition distribution $P_{\theta_s}(s_{t+1}|s_t, a_t)$, we have introduced the subscript $s$ for state transition distribution parameters $\theta_s$ (and for the function space view representation $f_s(\cdot)$ respectively). The corresponding graphical model for the partially observable setting is depicted in Figure~\ref{fig:vi_mbrl_po} and recovers the graphical model for the fully observable setting from Figure~\ref{fig:vi_mbrl} when ignoring the nodes for the observations $o_t$ and the observation distribution parameter $\theta_o$.

Since $\theta_s$, $\theta_o$ and $s_1, ..., s_T$ are all latent variables, we specify the prior of the generative model as $p_\gamma(\theta_s) p_\gamma(\theta_o) p(s_1, ..., s_T | s_0, a_0, ..., a_{T-1}, \theta_s)$ where $\gamma$ refers to the entirety of all generative parameters and where we assume a factorized form as in earlier parts of this tutorial. The state prior is then defined as $p(s_1, ..., s_T | s_0, a_0, ..., a_{T-1}, \theta_s) = \prod_{t=1}^{T} P_{\theta_s}(s_{t+1}|s_t, a_t)$, and the likelihood as $p(o_1,...,o_T|s_1, ..., s_T,\theta_o) = \prod_{t=1}^{T} P_{\theta_o}(o_t|s_t)$. Note that the likelihood is only conditioned on latent states $s_1$ up to $s_T$ and the observation distribution parameters $\theta_o$, but not on $s_0$ and the actions $a_0, ..., a_{T-1}$ as well as $\theta_s$, due to the conditional independence assumptions in the graphical model. Under an approximate posterior of the form $q_\psi(\theta_s) q_\psi(\theta_o) q_\psi(s_1, ..., s_T)$ where $q_\psi(s_1, ..., s_T) = \prod_{t=1}^T q_\psi(s_t)$, we finally arrive at the following $\ELBO$:
\begin{eqnarray}
\ELBO(\gamma, \psi) &=& \int \int ... \int q_{\psi}(\theta_o) q_\psi(s_1, ..., s_T) \ln p(o_1,...,o_T|s_1, ..., s_T,\theta_o) \; \textrm{d} s_{T} ... \; \textrm{d} s_{1} \; \textrm{d} \theta_o \nonumber \\
&& - \; \KL \Big( q_{\psi}(\theta_s) q_\psi(s_1, ..., s_T) \Big| \Big| p_{\gamma}(\theta_s) p(s_1, ..., s_T | s_0, a_0, ..., a_{T-1}, \theta_s)  \Big) \nonumber \\
&& - \; \KL \Big( q_\psi(\theta_o) \Big| \Big| p_\gamma(\theta_o)  \Big) \label{eq:vi_mbrl_po} \\
&=& \sum_{t=1}^{T} \int \int q_\psi(\theta_o) q_\psi(s_{t}) \ln P_{\theta_o}(o_{t}|s_{t}) \; \textrm{d} s_{t} \; \textrm{d} \theta_o \nonumber \\
&& - \sum_{t=1}^{T} \int \int q_{\psi}(\theta_s) q_\psi(s_{t-1}) \KL \Big( q_\psi(s_{t})) \Big| \Big| P_{\theta_s}(s_{t} | s_{t-1}, a_{t-1})  \Big) \; \mathrm{d} s_{t-1} \; \mathrm{d} \theta_s \nonumber \\
&& - \; \KL \Big( q_\psi(\theta_s) \Big| \Big| p_\gamma(\theta_s)  \Big) - \KL \Big( q_\psi(\theta_o) \Big| \Big| p_\gamma(\theta_o)  \Big). \label{eq:vi_mbrl_po_2}
\end{eqnarray}
Note that the second line in Equation~\eqref{eq:vi_mbrl_po} expresses the $\KL$ of the joint distribution over $\theta_s$ and $s_1, ..., s_T$ between the approximate posterior and the prior (because the state transition distribution $P_{\theta_s}(s_{t+1}|s_t,a_t)$ depends on $\theta_s$). This term then decomposes into two terms in Equation~\eqref{eq:vi_mbrl_po_2} (in line two and three). On a more subtle note, the second line of Equation~\eqref{eq:vi_mbrl_po_2} implicitly specifies $q_\psi(s_{0})$ for notational convenience (because $t$ starts at $1$), which is defined as a Dirac delta function over the given initial state $s_0$ (and is hence not part of the variational model).

Predicting future observations $o_1^\star,o_2^\star,...,o_{T^\star}^\star$ for a given initial state $s_0^\star$ and a given action sequence $a_0^\star,a_1^\star,a_2^\star,...,a_{T^\star-1}^\star$ is then accomplished via the following:
\begin{eqnarray}
p(o_1^\star, ..., o_{T^\star}^\star | s_0^\star, a_0^\star, ..., a_{T^\star-1}^\star) = \int \int \int ... \int  \prod_{t=0}^{T^\star-1} P_{\theta_o}(o_{t+1}^\star | s_{t+1}^\star) P_{\theta_s}(s_{t+1}^\star|s_t^\star,a_t^\star) \; \textrm{d} s_{T^\star} ... \; \textrm{d} s_{1}  \nonumber \\
 q_\psi(\theta_o) q_\psi(\theta_s) \; \mathrm{d} \theta_o \; \mathrm{d} \theta_s . \; \; \; \; \; \; \; \; \; \; \; \; \; \; \; \; \; \; \; \; \; \; \; \; \; \; \; \; \; \; \; \; \; \; \; \; \; \;
\label{eq:state_prediction_po}
\end{eqnarray}
In order to sample an observation sequence, one first has to sample a state-transition model $\theta_s$ and an observation model $\theta_o$ from the approximate posteriors $q_\psi(\theta_s)$ and $q_\psi(\theta_o)$ respectively. Subsequently, one can ``unroll'' the trajectory starting from $s_0^\star$ and $a_0^\star$ by generating $s_1^\star \sim P_{\theta_s}(s_1^\star|s_0^\star,a_0^\star)$ and $o_1^\star \sim P_{\theta_o}(o_1^\star|s_1^\star)$, and repeating this procedure until one arrives at $s_{T^\star}^\star$ and $o_{T^\star}^\star$. Note that, in accordance with previous sections, the approximate state posterior $q_\psi(s_1, ..., s_T)$ is not used for prediction, but only used during training and ``thrown away'' afterwards.

In practice, it might be more important to infer latent states $s_t$ on the fly at each time step~$t$ as the agent interacts with its environment, rather than generating new observation sequences. The reason why this might come in handy is that the agent could then use a state-conditioned policy $\pi(a_t|s_t)$ both for virtual planning but also for acting. One way of achieving this is to simply parameterize the approximate posterior over latent states as $q_\psi(s_t|o_t)$ in an amortized fashion that maps observations~$o_t$ probabilistically to states~$s_t$. While this has the advantage of enabling inference at act time, it ignores a lot of useful information at train time (since all the other observations and all actions carry information for inferring $s_t$). It is hence advisable to specify the approximate state posterior as $q_\psi(s_t|s_0,o_1,...,o_T,a_0,...,a_{T-1})$ with an explicit dependence on past and future time steps, and to learn a separate inference model for acting e.g.\ of the form $q_\chi(s_t|s_0,o_1,...,o_t,a_0,...,a_{t-1})$ that, importantly, does not require future time steps to infer the current latent state $s_t$ at time step $t$. Here, $\chi$ indicates that the parameters for this inference model are not part of the $\ELBO$ objective and are hence not contained in the variational parameters~$\psi$, but need to be identified through another objective (for example via filtering). How to precisely model variational posteriors in a POMDP setting is however outside the scope of this tutorial and we refer the reader to~\cite{Tschiatschek2018} and \cite{Eleftheriadis2017} as examples for neural network and Gaussian process models respectively.

\section{Summary}
\label{sec:conclusion}

The aim of this manuscript is to demonstrate the connection between VI and RL, both in model-free and model-based scenarios. Section~\ref{sec:vi} therefore starts by providing an overview of general VI. Sections~\ref{sec:contextual_bandits} and~\ref{sec:rl} show how to recover the RL objective to maximize expected future cumulative reward under a soft policy constraint from an $\ELBO$ formulation, for a non-sequential and a sequential setting respectively. In Sections~\ref{sec:mbrl} and~\ref{sec:mbrl_po}, we also show how to learn environment models with VI for sequential decision-making problems, both when the environment is fully and partially observable.




\bibliography{gp_tutorial}  

\end{document}